\documentclass{article}
\usepackage{url}
\usepackage[pdftex]{graphicx}
\usepackage{amsmath}
\usepackage{booktabs}
\usepackage{amsfonts}
\usepackage{float}
\usepackage{natbib}
\usepackage{multirow}
\usepackage{authblk}  

\title{STRuCT-LLM: Unifying Tabular and Graph Reasoning with Reinforcement Learning for Semantic Parsing}

\author{
    \textbf{Josefa Lia Stoisser}\textsuperscript{*}, \textbf{Marc Boubnovski Martell}\textsuperscript{*}, \textbf{Lawrence Phillips}, \textbf{Casper Hansen}, \textbf{Julien Fauqueur} \\
    Novo Nordisk \\
    \texttt{OFSR@novonordisk.com, MBVK@novonordisk.com, LAWRENCE@novonordisk.com, CASPER@novonordisk.com, JLZF@novonordisk.com} \\
}
\date{June 2025} 

\begin{document}

\maketitle

\begin{abstract}
We propose STRuCT-LLM, a unified framework for training large language models (LLMs) to perform structured reasoning over both relational and graph-structured data. Our approach jointly optimizes Text-to-SQL and Text-to-Cypher tasks using reinforcement learning (RL) combined with Chain-of-Thought (CoT) supervision. To support fine-grained optimization in graph-based parsing, we introduce a topology-aware reward function based on graph edit distance. Unlike prior work that treats relational and graph formalisms in isolation, STRuCT-LLM leverages shared abstractions between SQL and Cypher to induce cross-formalism transfer, enabling SQL training to improve Cypher performance and vice versa—even without shared schemas. Our largest model (QwQ-32B) achieves substantial relative improvements across tasks: on semantic parsing, Spider improves by 13.5\% and Text2Cypher by 73.1\%. The model also demonstrates strong zero-shot generalization, improving performance on downstream tabular QA (TableBench: 8.5\%) and knowledge graph QA (CR-LT-KGQA: 1.7\%) without any QA-specific supervision. These results demonstrate both the effectiveness of executable queries as scaffolds for structured reasoning and the synergistic benefits of jointly training on SQL and Cypher (code available at \url{https://github.com/bouv/STRuCT-LLM}).
\end{abstract}

\noindent\textsuperscript{*} Equal contribution. Listing order is random.

\section{Introduction}

Large language models (LLMs) demonstrate impressive fluency in open-domain generation but often falter on structured reasoning tasks involving tables and graphs \cite{jiang2023structgpt, guo2023gpt4graph}. Structured reasoning requires models to ground entities, compose symbolic constraints, and follow logical paths—skills crucial for interacting with real-world data systems such as relational databases and knowledge graphs (KGs) \cite{li2023can, pourreza2023din}. We view executable semantic parsing—specifically, Text-to-SQL and Text-to-Cypher—as a gateway to this broader capability~\cite{sui-etal-2024-tap4llm, ozsoy2024text2cypher}. While Text-to-SQL is well-studied, Text-to-Cypher remains underexplored, offering a valuable testbed for graph reasoning. These tasks provide a concrete, supervised setting in which models must map natural language to compositional, structured programs over relational (SQL) and graph (Cypher) formalisms. While typically treated in isolation, both involve shared abstractions—schema grounding, filtering, joins, and path composition—making them fertile ground for studying cross-formalism transfer.

Our hypothesis is that reinforcement learning (RL) over both Text-to-SQL and Text-to-Cypher tasks enables language models to acquire complementary reasoning skills—relational operations from SQL and graph traversal from Cypher. This synergy should enhance performance on semantic parsing and downstream knowledge graph QA. By training a unified model and leveraging structural transfer across query formats, the model develops structure-aware reasoning capabilities that generalize across both table and graph-structured data.

\begin{figure}
  \centering
  \includegraphics[width=0.8\linewidth]{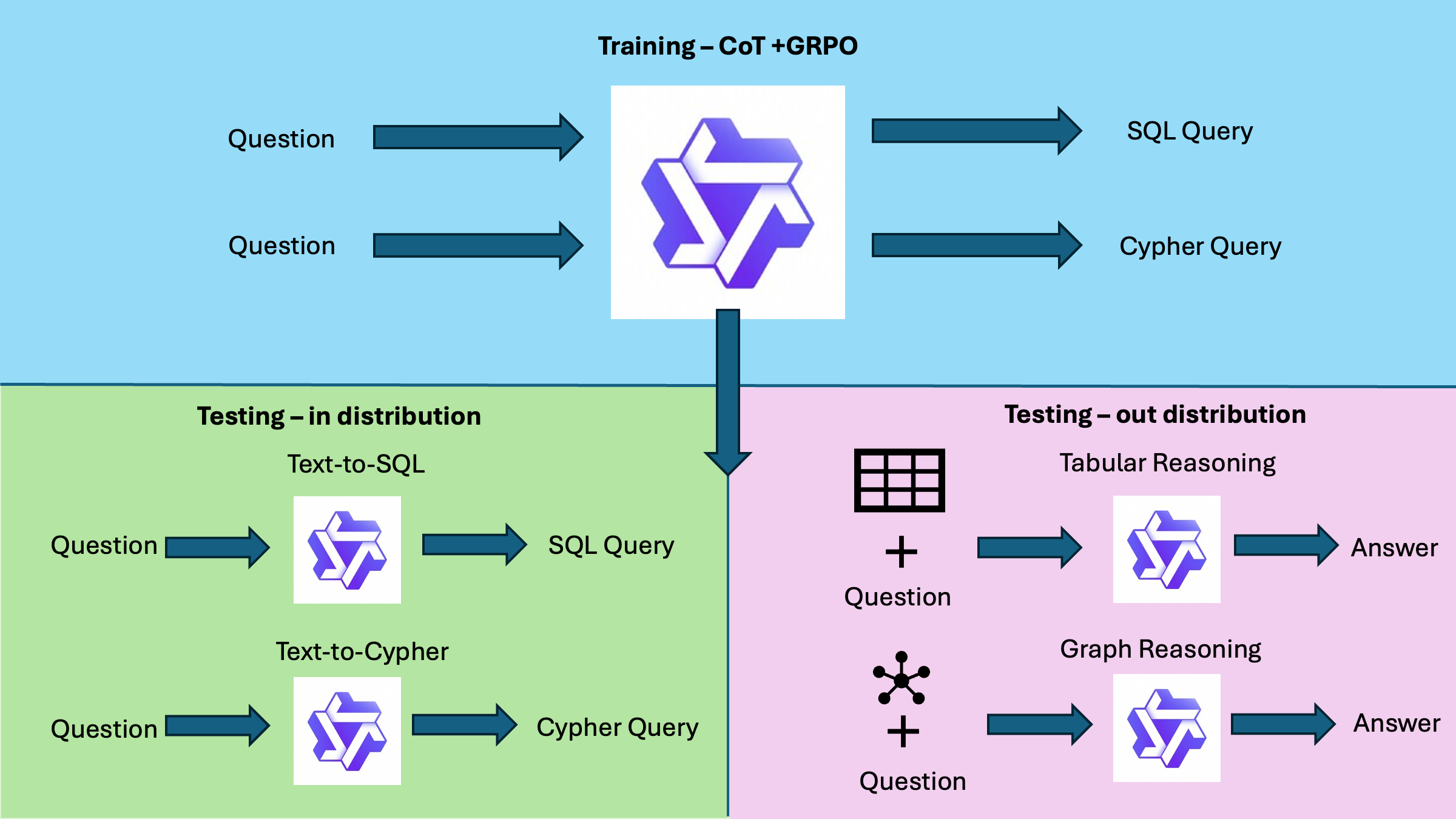}
  \caption{\textbf{Overview of our unified training and evaluation setup.} Joint training on Text-to-SQL and Text-to-Cypher using Chain-of-Thought supervision and RL (Task 1). Evaluation covers generalization to SQL/Cypher (Task 2), zero-shot transfer to tabular (Task 3), and graph QA (Task 4), testing for transferable reasoning from executable semantic parsing.}
  \label{fig:sample}
\end{figure}

To explore this hypothesis, we adopt a unified training setup combining Chain-of-Thought (CoT) supervision and RL, as seen in Figure \ref{fig:sample}. We prompt instruction-tuned LLMs (Qwen2.5-1.5B, 14B, QwQ-32B, Qwen3-14B) \cite{yang2024qwen2} with decomposed reasoning steps for both SQL and Cypher, and apply Group Relative Policy Optimization (GRPO) ~\cite{shao2024deepseekmath} using execution, structural, and syntactical feedback as a reward. For Cypher, we introduce a topology-aware reward that measures correctness over nodes, edges, and relationships—enabling RL optimization in the graph domain for the first time. Our joint training regime interleaves Text-to-SQL and Text-to-Cypher data across both supervised and RL stages, encouraging structural transfer and mitigating task-specific overfitting.

We find that jointly training on SQL and Cypher leads to mutual gains on both formats, leveraging the complementary inductive biases of relational and graph queries. Moreover, the resulting models exhibit promising zero-shot generalization to broader table and graph QA tasks (CRT-QA, Tablebench, CR-LT-KGQA), despite having never been trained on them directly. These results suggest that executable queries are not merely a target—they are a vehicle for teaching LLMs how to reason over structured data.

\subsection*{Contributions}

\begin{itemize}
    \item \textbf{Joint Training:}  We interleave SQL and Cypher data during both supervised and RL stages, enabling cross-formalism structural transfer and reducing task-specific overfitting.
    
    \item \textbf{Topology-Aware Reward:} We propose a novel subgraph-based reward function for Cypher semantic parsing that captures correctness over nodes, edges, and connectivity—enabling fine-grained, execution-based optimization in the graph domain.

    \item \textbf{CoT + RL Synergy:}  We integrate CoT supervision with RL to promote compositional intermediate reasoning steps across both query formats.
    
    \item \textbf{Cross-formalism Generalization:} Jointly trained models consistently outperform SQL-only and Cypher-only models on semantic parsing tasks, effectively leveraging complementary reasoning signals. Additionally, the models generalize to a completely unseen query formalism MQL, demonstrating strong cross-formalism adaptability.
    
    \item \textbf{Zero-Shot Transfer to QA:} Without explicit QA supervision, our models show promising improvement on zero-shot performance on table and KG QA benchmarks ~\cite{wu2025tablebench,zhang2023crt, guo2024cr}, suggesting the emergence of transferable structured reasoning capabilities.

\end{itemize}

\section{Related Work}

Semantic parsing for relational databases has evolved through benchmarks such as Spider~\cite{yu2018spider}  and Bird \cite{li2023can}, alongside transformer-based models that incorporate schema linking ~\cite{liu2024solid,katsogiannis2023survey} and execution-guided decoding ~\cite{huang2024datacentric}. RL has also been explored in this setting, optimizing for execution accuracy ~\cite{yang-etal-2024-synthesizing}  or syntactic validity ~\cite{ma2025sql}. While most prior work treats Text-to-SQL in isolation, recent efforts explore structural transfer across formalisms such as graphs  ~\cite{xie-etal-2022-unifiedskg}. SM3-Text-to-Query \cite{sivasubramaniam2024smtexttoquery}  introduced a synthetic dataset with 40k samples spanning SQL, MQL, Cypher, and SPARQL, enabling evaluation across four distinct query languages.

Parsing Cypher queries introduces unique challenges due to dynamic graph topologies and traversal semantics \cite{ozsoy2025enhancing,  ozsoy2025text2cypher}. Progress had been limited by scarce training data \cite{zou2024q2cypher}, but synthetic datasets like MedT2C ~\cite{zhong-etal-2025-synthet2c} and SynthCypher ~\cite{tiwari2025auto} have expanded coverage. To generate SynthCypher, ~\cite{tiwari2025auto} modeled each relational table as a node and each foreign key as an edge, demonstrating the structural similarity between relational and graph schemas. Recently, a dataset of Text-to-Cypher mappings with over 44k examples was released, alongside a fine-tuned model for Cypher generation \cite{ozsoy2024text2cypher}. Our work contributes a topology-aware reward function aligned with subgraph connectivity, enabling RL for Cypher based on graph-structured feedback.

Parallel efforts in SPARQL parsing and MongoDB Query Language (MQL) have emphasized topological structure in graph queries  \cite{xu-etal-2023-fine, lu2025bridging}. Neural SPARQL parsers ~\cite{yin2021neural} encode graph patterns to generate triple-based queries over ontological KGs such as Freebase and DBpedia. While distinct in formalism, these approaches demonstrate the value of structure-aware modeling—a principle we extend to property graph parsing via connectivity-sensitive rewards.

Recent advancements in table-based question answering Torr \cite{ashury2025mighty} and(TableQA) have been propelled by datasets such as WikiSQL, Spider, and OTTQA, which enable multi-hop reasoning \cite{stoisser2025sparks, zhang2023crt}. However, many existing benchmarks focus on fact retrieval and overlook commonsense reasoning, which is essential for nuanced real-world applications \cite{fang2024large}. LLMs excel in structured reasoning; they are prone to hallucinations, especially in long-tail knowledge settings \cite{guo2024cr}. To mitigate this, KGs provide structured supervision for grounding LLMs and improving query precision \cite{zhang2024knowgpt}.

Although Knowledge graph QA (KGQA) research has advanced, it often neglects commonsense reasoning and long-tail entities. To address this gap, we introduce CR-LT KGQA \cite{guo2024cr}, a novel dataset emphasizing commonsense reasoning and rare-entity coverage, enhancing the applicability of KGQA in tandem with TableQA progress.

Recent multi-task pretraining and adapter-based frameworks have shown the benefits of shared supervision across symbolic tasks. UnifiedSKG ~\cite{xie-etal-2022-unifiedskg} and SmBoP ~\cite{rubin-berant-2021-smbop} learn from heterogeneous sources including SQL, SPARQL, and KGQA. While architecturally distinct, these efforts echo our motivation: to induce generalizable structured reasoning. Our contribution lies in combining multi-formalism supervision with CoT prompting ~\cite{wei2022chain} and RL \cite{guo2025deepseek} tailored to execution and graph topology.

\section{Methodology}

\subsection{Problem Formulation}
In structured query languages such as SQL and Cypher, the query space, \(\mathcal{Q}\), can be defined as a sequence of transformations applied to a predefined schema of data. Let \(D\) denote the initial dataset, consisting of tables \(T = \{T_1, T_2, \ldots, T_p\}\) and the database schema \(S\) in SQL, or nodes \(N = \{N_1, N_2, \ldots, N_q\}\) and relationships \(R = \{R_1, R_2, \ldots, R_s\}\) in Cypher. The query space \(\mathcal{Q}\) is generated through chains of operations \(\mathcal{O}\) on the given data \(D\), guided by schema \(S\), and expressed as:

\[
\mathcal{Q} = \{Q \mid Q: D \rightarrow D', \: D' \subseteq \mathcal{R}_{\text{Result}}\}
\]

where \(Q\) represents a composite function consisting of operations \(\mathcal{O}_i\):

\[
Q(D) = \mathcal{O}_n (\mathcal{O}_{n-1}(\ldots \mathcal{O}_2(\mathcal{O}_1(D)) \ldots))
\]

Each operation \(\mathcal{O}_i\) can be a selection, projection, join , pattern match, etc., transforming \(D\) sequentially. The result space \(\mathcal{R}_{\text{Result}}\) represents valid query outcomes subject to language grammar and data model constraints. Semantic parsing is defined as the task of transforming a natural language input \(X\) into a target query \(Q\) within the space \(\mathcal{Q}\), such that \(Q\) accurately retrieves the specific information from \(D\) requested by \(X\).

To enhance semantic parsing across these paradigms, our approach jointly trains models on two semantic parsing tasks: Text-to-SQL, utilizing SQLite as the query language, and Text-to-Cypher. This methodology involves two key stages: (1) supervised fine-tuning on synthetically generated CoT traces, and (2) RL employing multiple reward components. By leveraging the shared formulation and the synergistic relationship between SQL and Cypher query generation tasks, our strategy aims to deepen the model's understanding of structured data manipulation.

\subsection{Datasets}
\textbf{Supervised Fine-tuning Training Datasets}

Building on the efficacy of CoT supervised fine-tuning for semantic parsing~\cite{ma2025sql}, we generate synthetic CoT traces for Text-to-SQL and Text-to-Cypher tasks, leveraging Large Language Models through a structured prompting pipeline as in \citet{Boubnovski2025scalable}. This process incorporates a summarization step and involves prompting the o3-mini model to provide query answers with natural language reasoning, verified by a second model for answer and reasoning coherence (details in Appendix~\ref{app:prompts}). We apply this methodology across 24 Text-to-SQL datasets\footnote{\url{https://huggingface.co/datasets/Clinton/Text-to-sql-v1}} and the Neo4j Text-to-Cypher dataset~\cite{ozsoy2024text2cypher}, see Appendix ~\ref{app:data} for details. This diverse dataset foundation supports an effective cold start for GRPO training. For experiments, we construct 3,500 data points per task and use only half of them for joint training.


\textbf{GRPO Training Datasets}

\underline{SQL Dataset:} We leverage the BIRD benchmark~\cite{li2023can} for its complex multi-step queries, ideal for GRPO's nuanced reward optimization. We exclude BIRD's evidence information to ensure generalization to other datasets without dependency on supplementary data.

\underline{Cypher Dataset:} Given the limited availability of non-synthetic Cypher datasets, we utilize a distinct split of the Neo4j Text-to-Cypher dataset~\cite{ozsoy2024text2cypher}, ensuring no overlap with data used for supervised fine-tuning. 

For both tasks, we compile 3,500 data points to facilitate fair comparisons between single-language and joint training scenarios. Single-task training includes 7,000 SQL examples, while joint training allocates 3,500 examples to each SQL and Cypher, maintaining balanced representation across tasks.

\subsection{Training Framework}

\subsubsection{Reinforcement Learning}

We employ GRPO, an RL method originally introduced in \citet{shao2024deepseekmath}. This approach enhances traditional RL by comparing multiple outputs for the same input and assigning relative rewards, enabling more nuanced feedback for complex reasoning tasks.

Formally, for a given natural language question $q$ and its associated database schema, the model generates a set of $G$ candidate SQL or Cypher queries $\{o_1, o_2, \ldots, o_G\}$. Each candidate is evaluated using a task-specific reward function, and the relative advantage $A_i$ is computed for each output. The optimization objective is given by:

\begin{align}
    J_{GRPO}(\Theta) = & \mathbb{E} \left[ \frac{1}{G} \sum_{i=1}^{G} \min \left( \frac{\pi_\theta(o_i | q)}{\pi_{\theta_{\text{old}}}(o_i | q)} A_i, \text{clip} \left( \frac{\pi_\theta(o_i | q)}{\pi_{\theta_{\text{old}}}(o_i | q)}, 1 - \epsilon, 1 + \epsilon \right) A_i \right) \right] \nonumber  - \beta D_{KL}(\pi_\theta || \pi_{\text{ref}})
\end{align}

Here, $\pi_\theta$ represents the current policy, $\pi_{\theta_{\text{old}}}$ denotes the policy before the update, and $\pi_{\text{ref}}$ is a frozen reference policy used for regularization. The hyperparameters $\epsilon$ and $\beta$ control the clipping threshold and divergence penalty, respectively.

\subsubsection{Reward Design} \label{sec:reward_design}
The primary objective in Text-to-SQL and Text-to-Cypher tasks is to generate queries that execute to the correct results. Traditional binary execution rewards are ineffective for learning from nearly correct predictions due to their lack of gradient information. To address this limitation, we implement continuous rewards that guide the model towards generating accurate and structurally consistent queries, similarly to \cite{stoisser2025sparks}.

\begin{enumerate}
    \item \textbf{LLM Judge Reward with Classes:} For both SQL and Cypher query generation, this reward provides feedback on the overall quality of the response. Using OpenAI's o3-mini model \cite{jaech2024openai} as judge, we evaluate the generated query by classifying it into ordinal categories that capture query quality, inspired by \citet{xin2024deepseek}. The specific prompts and classification categories are detailed in Appendix~\ref{app:prompts}.
    
    \item \textbf{String Matching Reward:} Offering a continuous approach to measure query similarity, this reward computes the longest contiguous matching subsequence between predicted and gold query strings for both SQL and Cypher. By quantifying similarity, it encourages syntactical accuracy.
    
    \item \textbf{Structural Consistency Rewards:} 
    Because queries inherently possess structure and multiple syntactically different queries may produce identical results, structured rewards—adapted to each query language—are essential for ensuring semantic fidelity.
\end{enumerate}

For structural consistency in SQL queries, we compute Component-Level Matching Reward, which calculates the exact match between predicted and ground truth SQL components (e.g., \texttt{SELECT}, \texttt{WHERE}), as used in \cite{stoisser2025sparks} and \cite{NGUYEN2025100135} - using the F1 score. This reward is particularly suited for SQL's compositional nature, where queries often involve multiple nested operations and complex combinations of components. By evaluating each component separately, we can provide granular feedback on the model's ability to correctly generate each operation.

For Cypher queries, we design a novel Graph Edit Distance-based Reward. Cypher queries primarily focus on pattern matching within graph structures, and we design our reward to emphasize this topological understanding rather than operational correctness. We construct property graph representations of the subgraphs targeted by the gold and predicted Cypher queries, then compute the Graph Edit Distance (GED) between them, which quantifies the minimal set of edit operations required to transform a predicted subgraph into its gold-standard counterpart. The reward is computed as:
\begin{equation}
    R_{\text{GED}} = 1 - \frac{\text{GED}}{\max(\text{size}_1, \text{size}_2)},
\end{equation}
where $\text{size}_1$ and $\text{size}_2$ are the total counts of nodes and edges in each graph, ensuring a normalized assessment of fidelity with respect to the target structure.

\subsubsection{Joint Training Approach}

We propose a joint training strategy combining Text-to-SQL and Text-to-Cypher objectives by leveraging shared principles in structured data manipulation and logical reasoning. This paradigm is motivated by several factors. First, it leverages shared structural priors to help the model generalize across data operations. Second, training on diverse examples induces a regularization effect, reducing overfitting. Finally, it improves sample efficiency through potential transfer learning from query similarities.

We interleave Text-to-SQL and Text-to-Cypher examples for the training. For GRPO, we employ task-specific reward combinations for SQL ($w_1R_{\text{judge}} + w_2R_{\text{string}} + w_3R_{\text{component}}$) and Cypher ($w_1R_{\text{judge}} + w_2R_{\text{string}} + w_3R_{\text{GED}}$) queries. Here, $R_{\text{judge}}$ and $R_{\text{string}}$ represent LLM judge and string matching rewards, while $R_{\text{component}}$ and $R_{\text{GED}}$ denote structural similarity rewards detailed in Section~\ref{sec:reward_design}. We set all weights equally ($w_1=w_2=w_3=1$) without tuning; exploring adaptive weights remains future work.


\section{Experiments}

\subsection{Dataset} \label{sec:dataset}
We evaluate our model based on two primary criteria: (1) its performance on semantic parsing tasks, specifically through its generalization to novel databases, is assessed using Spider \cite{yu2018spider}, EHRSQL \cite{lee2022ehrsql} and BIRD minidev \footnote{\url{https://GitHub.com/bird-bench/mini_dev}} for Text-to-SQL and Text2Cypher \cite{ozsoy2024text2cypher} for Text-to-Cypher; and (2) its transfer capability to out-of-distribution tasks in the Table/KG Question Answering, which reflects its induced reasoning capabilities, is evaluated using CRT-QA \cite{zhang2023crt} for tabular and CR-LT KGQA \cite{guo2024cr} for KG QA. Additionally, we leverage the SM3 dataset \cite{sivasubramaniam2024smtexttoquery}, which provides semantically equivalent queries across SQL, Cypher, and MQL, enabling direct comparison of model performance across different query languages. Further details can be found in Appendix~\ref{app:data}.

\subsection{Experimental Setup}

In our experiments, we use both non-reasoning models (Qwen2.5-1.5B, Qwen2.5-14B) and reasoning models (QwQ-32B, Qwen3-14B) as baselines. We utilize VERL\footnote{\url{https://GitHub.com/volcengine/verl}} for training the 14B and 32B models. To enhance efficiency, Unsloth\footnote{\url{https://GitHub.com/unslothai/unsloth}} is employed, limiting the training to QLora Adapters with a LoRA rank of 16 for all 1.5B models. For Supervised Fine-Tuning, we adopt a learning rate of $1e^{-6}$ for larger models (14B and 32B) and $1e^{-4}$ for smaller models. The batch size is set to 512 for big models and 8 for small models. For GRPO, consistent with \citet{pourreza2025reasoning}, a constant learning rate scheduler is applied. Training is conducted for one epoch with a learning rate of $1e^{-6}$ for large models and $1e^{-5}$ for small models. Additional hyperparameters for GRPO include a number of generations $G=6$, with batch sizes of 1024 for large models and 32 for small models, using the \texttt{PagedAdamW8bit} optimizer. For robust evaluation, we sample 200 datapoints from each dataset five times and report mean scores and standard deviations. For training we use 32 H100 GPUs for 60 h for big models. Additional hyperparameters are available in our GitHub repository \footnote{\url{https://github.com/bouv/STRuCT-LLM}}.

\subsection{Ablation Studies} \label{sec:ablations}
We conducted comprehensive ablation studies to evaluate: (1) the impact of different training configurations (SQL-only, Cypher-only, and joint training), and (2) the contribution of each reward component in our GRPO framework. Our analysis employed Qwen2.5-1.5B with results presented in Table~\ref{tab:appendix_ablations}.

\paragraph{Training Configuration Analysis}
The results reveal a positive transfer between SQL and Cypher even with single-domain training—models trained solely on Cypher show gains of 0.4-0.8\% on Spider, while SQL-trained variants demonstrate improvements of 0.9-2\% on Text2Cypher. Notably, joint training exhibits clear synergy: Both-111 outperforms Cypher-111 by 0.4\% on Text2Cypher tasks while surpassing SQL-111 by 1.7\% on Spider, demonstrating effective bidirectional knowledge transfer across domains.

\paragraph{Reward Component Analysis}
We evaluate the impact of different reward components (string matching, LLM-based, and structural) through systematic ablation, denoted by 1/0 in model names. The full reward configuration (111) consistently achieves optimal performance, highlighting the complementary nature of these components. Ablation experiments reveal the distinct role of each component: removing string matching rewards (011) causes notable performance drops, especially in semantic parsing tasks, while disabling LLM-based rewards (101) leads to moderate degradation. The structural reward proves particularly crucial—its removal (110) significantly impacts performance, especially on QA tasks where structural comprehension is essential.

Overall, these analyses demonstrate the synergistic benefits of our multi-component reward structure and joint training approach. Additional ablation details are provided in Appendix~\ref{tab:appendix_ablations}.

\subsection{In-Context Performance}
Table~\ref{tab:in_context} presents the performance on semantic parsing tasks. Training on a single executable language (SQL or Cypher) induces \textit{slight positive transfer} to the other: for instance, training Qwen3-14B on Cypher alone improves its SQL performance on Spider (EXE: 74.2 vs. 73.7), and vice versa for Text2Cypher. This cross-task improvement suggests structural similarities—such as schema grounding and compositional filtering—shared between the two tasks.

Motivated by this, we train models on both SQL and Cypher jointly. The resulting models demonstrate strong \textit{synergy}, consistently outperforming single-task variants across all base models. For example, Qwen2.5-14B-trained-Both outperforms both SQL-only and Cypher-only versions on all tasks.

Notably, our QwQ-32B-trained-Both model \textit{approaches or exceeds} the performance of proprietary instruction-tuned models such as o3 and o3 mini, particularly on Text2Cypher. This suggests that dual-format supervision is a powerful training signal, enabling open models to compete with the best closed models on structural parsing tasks.

These results are on main benchmarks within their respective domains; however, they are not comparable to each other directly. Hence, in Appendix \ref{app:sm3}, we include additional results on the SM3-Text-to-Query Dataset \cite{sivasubramaniam2024smtexttoquery}, which provides semantically equivalent queries across SQL, Cypher, and MongoDB Query Language (MQL). This dataset allows us to evaluate how well our training approach generalizes to unseen query languages.

We observe that base models perform better on SQL compared to Cypher and show significantly weaker performance on MQL, reflecting the data availability during pretraining (SQL being the most abundant and MQL the scarcest). On the other hand, our models achieve superior performance in Cypher compared to SQL, potentially due to Cypher's relative simplicity or the greater effectiveness of our structural reward mechanism in tackling graph-based tasks.

\begin{table}[t]
\centering
\small
\caption{\textbf{ In-context learning performance on Text-to-SQL (Spider, EHRSQL, BIRD minidev with evidence) and Text-to-Cypher (Text2Cypher) tasks}.
Metrics include exact match (EM), execution accuracy (EXE), BLEU score and execution-based F1 score ($F1_{exe}$, per \cite{lee2022ehrsql} ). Models are labeled as ‘trained-SQL’, ‘trained-Cypher’, or ‘trained-Both’ based on fine-tuning. Top two results per dataset are bolded.}
\label{tab:in_context}
\begin{tabular}{lcccccc}
\toprule
\multirow{2}{*}{\textbf{Model}} & \multicolumn{2}{c}{\textbf{Text2Cypher}} & \textbf{BIRD} & \textbf{Spider} &  \textbf{EHRSQL} \\
\cmidrule(lr){2-3} \cmidrule(lr){4-4} \cmidrule(lr){5-5} \cmidrule(lr){6-6}
& EM & BLEU & EXE & EXE & $F1_{exe}$ \\
\midrule
o3 & 4.0 {±0.6} & 25.0 {±2.2} & 54.0 {±4.5} & \textbf{76.8} {±1.6} &   \textbf{46.5} {±2.5} \\
o3-mini & 4.5 {±1.1} & 25.4 {±2.2} & 50.0 {±2.1} & 74.5 {±3.2} &   35.5 {±1.5} \\
\midrule
Qwen3-14B & 3.0 {±1.0} & 21.3 {±1.2} & 51.0 {±2.8} & 73.7 {±2.2} &   38.5 {±2.9} \\
Qwen3-14B-trained-SQL & 3.5 {±1.1} & 22.5 {±1.3} & 54.5 {±2.5} & \textbf{76.8} {±1.9} &  40.5 {±2.6} \\
Qwen3-14B-trained-Cypher & 5.5 {±1.2} & 26.8 {±1.4} & 52.2 {±2.6} & 74.4 {±1.8} &  39.5 {±2.7} \\
Qwen3-14B-trained-Both & 6.2 {±1.0} & \textbf{27.9} {±2.1} & \textbf{54.7} {±2.3} & 76.5 {±2.6} & \textbf{43.2} {±1.8} \\
\midrule
Qwen2.5-14B-instruct & 3.5 {±1.2} & 18.5 {±1.6} & 43.3 {±3.4} & 69.8 {±3.6} & 26.0 {±4.2} \\
Qwen2.5-14B-trained-SQL & 4.5 {±1.1} & 21.0 {±1.7} & 50.3 {±3.5} & 72.8 {±2.9}   & 32.3 {±3.6} \\
Qwen2.5-14B-trained-Cypher & 6.8 {±1.4} & 25.0 {±1.9} & 46.8 {±3.7} & 71.7 {±3.1} &  30.1 {±3.5} \\
Qwen2.5-14B-trained-Both & \textbf{7.3} {±1.3} & 25.2 {±2.0} & 50.4 {±3.2} & 74.5 {±2.6} &  33.6 {±3.8} \\
\midrule
QwQ-32B & 3.7 {±1.1} & 19.3 {±2.3} & 50.0 {±3.0} & 69.8 {±3.6}  & 35.0 {±2.6} \\
QwQ-32B-trained-Both & \textbf{12.0} {±1.5} & \textbf{33.4} {±3.6} & \textbf{55.3} {±3.0} & \textbf{79.2} {±4.0} &  {43.0} {±1.6} \\
\bottomrule
\end{tabular}
\end{table}


\subsection{Out-of-Context Performance}
Table~\ref{tab:out_context} presents the performance metrics on structural understanding tasks. Training on executable semantic parsing improves LLMs' ability to understand structured data more broadly, yielding better zero-shot performance on unseen question answering tasks. For example, both SQL- and Cypher-trained models outperform baselines on CRT-QA and CR-LT-KGQA, showing that execution-based reward helps models learn to reason over structured representations.

Moreover, joint training again provides strong synergy: Qwen2.5-14B-trained-Both, for example, achieves the best performance across Qwen2.5-14B models, with 62.5 EM on CRT-QA and 86.3 accuracy on CR-LT-KGQA—gains of 2.6 and 0.8 points over the best single-task variants. These results affirm that multi-format semantic supervision promotes generalization across structured modalities.

Additionally, the results in Appendix \ref{app:sm3} show that our models improve over baselines in few-shot settings and generalize effectively to the completely unseen structural language MQL. These findings reinforce the broad applicability of our approach to diverse structured reasoning tasks.

\begin{table}[t]
\centering
\small
\caption{\textbf{Out-of-context performance on structural question answering}. Results are reported on tabular (CRT-QA, TableBench) and knowledge graph QA (CR-LT-KGQA). Metrics include exact match (EM) and ROUGE. Models are labeled as ‘trained-SQL’, ‘trained-Cypher’, or ‘trained-Both’ based on fine-tuning. Top two results per dataset are bolded.
}
\label{tab:out_context}
\begin{tabular}{lcccccc}
\toprule
\multirow{2}{*}{\textbf{Model}} & \textbf{CRT-QA} & \multicolumn{2}{c}{\textbf{TableBench}} & \textbf{CR-LT KGQA} \\
\cmidrule(lr){2-2} \cmidrule(lr){3-4} \cmidrule(lr){5-5}
& EM & EM & Rouge & EM \\
\midrule
o3 & \textbf{61.5} {±2.3} & \textbf{68.0} {±4.4} & \textbf{73.1} {±3.5} & \textbf{92.0} {±0.7} \\
o3-mini & 45.2 {±4.0} & 64.8 {±4.4} & 68.6 {±3.8} & \textbf{91.7} {±2.3} \\
\midrule
Qwen3-14B & 53.3 {±3.6} & 66.3 {±3.5} & 72.1 {±2.7} & 82.8 {±2.0} \\
Qwen3-14B-trained-SQL & 56.2 {±3.2} & 67.8 {±3.7} & 72.8 {±2.9} & 83.7 {±1.9} \\
Qwen3-14B-trained-Cypher & 54.9 {±3.5} & 67.0 {±3.6} & 73.0 {±3.0} & 84.5 {±1.7} \\
Qwen3-14B-trained-Both & 55.6 {±3.1} & 67.0 {±3.8} & 73.0 {±3.1} & 84.2 {±1.8} \\
\midrule
Qwen2.5-14B-instruct & 55.5 {±1.3} & 55.5 {±4.7} & 58.3 {±4.1} & 80.2 {±1.2} \\
Qwen2.5-14B-trained-SQL & 59.9 {±1.6} & 62.9 {±3.9} & 62.0 {±3.5} & 82.5 {±1.3} \\
Qwen2.5-14B-trained-Cypher & 56.7 {±1.9} & 59.5 {±4.3} & 61.0 {±3.9} & 85.7 {±1.2} \\
Qwen2.5-14B-trained-Both & \textbf{62.5} {±1.7} & 63.8 {±3.5} & 65.5 {±3.9} & {86.3} {±1.4} \\
\midrule
QwQ-32B & 45.7 {±4.3} & 63.3 {±5.6} & 67.2 {±4.9} & 89.8 {±1.4} \\
QwQ-32B-trained-Both & 57.0 {±2.8} & \textbf{68.7} {±3.6} & \textbf{73.4} {±2.7} & 91.3 {±1.7} \\
\bottomrule
\end{tabular}
\end{table}

\section{Discussion}

Our findings reinforce the hypothesis that structured reasoning across distinct data modalities—relational tables and knowledge graphs—can be effectively unified through a shared training paradigm. By integrating Text-to-SQL and Text-to-Cypher supervision with RL, we demonstrate that large language models (LLMs) can acquire transferable inductive biases that transcend individual formalisms. The resulting STRuCT-LLM system displays not only improved accuracy on benchmark semantic parsing datasets but also exhibits zero-shot generalization to downstream QA tasks, underscoring the utility of executable queries as a scaffold for abstract reasoning.

A key insight from our study is the observed cross-formalism transfer. Despite the non-overlapping schemas and data sources between SQL and Cypher corpora, training on SQL alone induced gains on Cypher parsing, and vice versa. This suggests that LLMs, when trained with CoT supervision and topologically grounded rewards, can internalize structural abstractions—such as joins, paths, and filters—that generalize across both relational and graph-structured data. This transferability highlights the model’s emerging capacity for schema-agnostic reasoning, an encouraging signal for future multi-format semantic parsers.

Our topology-aware reward function for Cypher, built on graph edit distance and subgraph matching, proved crucial for enabling fine-grained optimization in the graph domain. Traditional execution-based rewards offer sparse or binary signals, whereas our design yields continuous, interpretable feedback, thereby addressing a longstanding bottleneck in Cypher-centric semantic parsing. More broadly, this reflects a growing shift in semantic parsing towards structurally sensitive learning objectives that move beyond mere execution correctness.

Joint training also acted as an effective regularizer. Compared to single-task training, our unified model exhibited improved robustness, with reduced overfitting to domain-specific patterns. This supports the view that structured multitask supervision not only encourages the learning of generalizable abstractions but also improves data efficiency, a desirable property when extending to lower-resource query languages or domains.

The zero-shot generalization to QA tasks—particularly CR-LT KGQA and CRT-QA—further confirms the downstream benefits of our approach. Notably, these tasks require reasoning over previously unseen schemas and long-tail entities, yet STRuCT-LLM achieved non-trivial gains without direct QA supervision. This finding aligns with recent observations that executable programs serve as a powerful proxy for latent logical structures, suggesting a promising avenue for future work in aligning natural language and symbolic representations via intermediate supervision.

In summary, our results demonstrate that blending relational and graph supervision with RL yields not only performance gains but also stronger generalization and reasoning capabilities. Future work includes: (i) support for additional query languages (SPARQL, MQL), (ii) interactive and schema-free settings with user feedback in reward design, and (iii) probing compositional generalization through specialized benchmarks to assess reasoning depth.

\section{Limitations}
While STRuCT-LLM demonstrates strong performance, it faces several limitations. The reliance on synthetic datasets—especially for Cypher—may affect generalization to real-world, noisy queries \cite{stoisser2025query}. Our supervised fine-tuning depends on LLM-generated CoT traces, which may introduce verification noise or bias. The reward functions, though structurally aware, do not account for query efficiency or execution cost. LLM-based judges used for reward computation may not provide fully reliable metrics. Additionally, the model assumes schema-aware inputs and has not been tested in interactive or schema-free settings. Finally, the combination of CoT and GRPO introduces notable computational overhead, potentially limiting accessibility for low-resource users. To strengthen the model's validity, additional evaluation across more diverse datasets and structural understanding tasks, such as table prediction, would be beneficial.

While we carefully controlled for direct overlap in our evaluation setup, some indirect overlap exists. The Spider evaluation benchmark may share some Wikipedia tables with our training data, though this represents a small fraction of the evaluation set. Similarly, our QA evaluation tasks may share some tables with the semantic parsing training data, although they test fundamentally different reasoning capabilities.

\section{Conclusion}

We introduced STRuCT-LLM, a unified semantic parsing framework that leverages RL and CoT supervision to jointly train on two fundamental query languages: SQL for relational databases and Cypher for graph databases. Our method demonstrates that training on just two structurally distinct formalisms can induce broad reasoning capabilities.

A key innovation enabling cross-formalism transfer is our topology-aware reward function for Cypher, which provides continuous, structure-sensitive feedback critical for RL optimization in graph domains. This allows models trained on SQL to improve at Cypher parsing and vice versa, despite differences in schema and format.

Future work includes: (i) support for additional query languages (SPARQL, MQL), (ii) interactive reasoning in agent systems with dynamic data sources, and (iii) probing compositional generalization under schema drift. Overall, STRuCT-LLM demonstrates that joint RL-based training on relational and graph query tasks yields models with enhanced structured reasoning capabilities, as evidenced by improved parsing and zero-shot QA generalization.

\bibliographystyle{plain} 
\bibliography{bib} 

\appendix

\section{Datasets} \label{app:data}
As detailed in Table~\ref{dataset-overview}, we utilize a variety of datasets for SQL and Cypher.

\subsection{SFT Training} \label{app:sft_data}

\textbf{SQL Dataset:} We utilize a collection of 24 publicly available Text-to-SQL datasets, selected for their moderate complexity to facilitate reliable chain-of-thought (CoT) trace generation. This collection is part of a dataset available on Hugging Face\footnote{\url{https://huggingface.co/datasets/Clinton/Text-to-sql-v1}}, which is a large-scale compilation of natural language to SQL examples spanning various domains. This benchmark includes 26 individual datasets that cover academic records, medical databases, entertainment metadata, government statistics, and more. Notable examples include WikiSQL \cite{zhong2017seq2sql}, ATIS \cite{hemphill1990atis}, Criteria2SQL \cite{fang2022combining}, SEDE \cite{hazoom2021text}, SQuALL \cite{shi2020potential}, and NVBench \cite{wang2023natural}, along with public domain table corpora such as IMDb, Yelp, and datasets on historical sports or wildfires. We exclude Spider \cite{yu2018spider} and EHRSQL \cite{lee2022ehrsql} from our training process, as they are designated for evaluation purposes, thus resulting in a total of 24 datasets used for training.

\textbf{Cypher Dataset:} The Text-to-Cypher dataset is derived from \cite{ozsoy2024text2cypher}, encompassing 16 standardized public datasets that represent a broad array of graph query scenarios. 

This strategic selection ensures comprehensive foundational development in both SQL and Cypher querying languages, facilitating an effective cold start for subsequent GRPO training.

\subsection{Evaluation}

We evaluate the model on two main criteria.  First, semantic parsing ability and generalization to novel databases through evaluation on various semantic parsing datasets. Second, we test transfer transfer capability to out-of-distribution Table/KG Question Answering tasks to demonstrate induced reasoning capabilities.

\noindent\textbf{Text-to-SQL Task:} For the Text-to-SQL Task, we use BIRD minidev\footnote{\url{https://GitHub.com/bird-bench/mini_dev}}, Spider \cite{yu2018spider}, EHRSQL \cite{lee2022ehrsql} and SM3 {sivasubramaniam2024smtexttoquery} benchmarks. EHRSQL is specifically chosen to assess generalization to entirely unseen databases; this is ensured by excluding all MIMIC \cite{johnson2016mimic} database-related questions from our training set since EHRSQL is based on MIMIC. Spider serves as a comprehensive benchmark for text-to-SQL capabilities.

\noindent\textbf{Text-to-Cypher Task:} Evaluation is conducted on the held-out test split of our training data \cite{ozsoy2024text2cypher} and on SM3 {sivasubramaniam2024smtexttoquery}.

\noindent\textbf{Tabular QA:} To assess transfer to table understanding, we evaluate on CRT QA \cite{zhang2023crt} and Tablebench \cite{wu2025tablebench}, which focus on direct reasoning over table data.

\noindent\textbf{KG QA:} We assess knowledge graph reasoning capabilities through evaluation on CR-LT KGQA \cite{guo2024cr}, which requires direct reasoning over graph structures.

\renewcommand{\thetable}{A\arabic{table}}
\setcounter{table}{0}

\begin{table}[t]
\caption{\textbf{Overview of Datasets used for Training and Evaluation}. The table shows the datasets used for different stages: Supervised Fine-Tuning (SFT), GRPO Training, and two types of evaluations.}
\label{dataset-overview}
\centering
\small
\begin{tabular}{@{}llll@{}}
\toprule
\textbf{Stage} & \textbf{Type} & \textbf{Table Task} & \textbf{Graph Task} \\
\midrule
\multirow{1}{*}{Training} & Semantic & Text-to-SQL Collection$^*$ & Neo4j Text-to-Cypher \\
&Parsing (SFT) & & \cite{ozsoy2024text2cypher} \\
\cmidrule(lr){2-4}
& Semantic & BIRD \cite{li2023can} & Neo4j Text-to-Cypher \\
&Parsing (GRPO) & & \cite{ozsoy2024text2cypher} \\
\midrule
\multirow{2}{*}{Evaluation} & Semantic & BIRD minidev$^\dagger$, & Neo4j Text-to-Cypher \\
& Parsing & EHRSQL \cite{lee2022ehrsql}, & \cite{ozsoy2024text2cypher} \\
& & Spider \cite{yu2018spider} & SM3 \cite{sivasubramaniam2024smtexttoquery} \\
& & SM3 \cite{sivasubramaniam2024smtexttoquery} & \\
\cmidrule(lr){2-4}
& Structural & CRT-QA \cite{zhang2023crt}, & CR-LT-KGQA \\
& QA & TableBench \cite{wu2025tablebench} & \cite{guo2024cr} \\
\bottomrule
\multicolumn{4}{@{}l@{}}{\scriptsize $^*$\url{https://huggingface.co/datasets/Clinton/Text-to-sql-v1}} \\
\multicolumn{4}{@{}l@{}}{\scriptsize $^\dagger$\url{https://GitHub.com/bird-bench/mini_dev}}
\end{tabular}
\end{table}

\section{Ablation Details} \label{app:ablations}
The full ablation results from training Qwen2.5-1.5B instruct are presented in table ~\ref{tab:appendix_ablations}. For their discussion we refer to Section \ref{sec:ablations}.

\renewcommand{\thetable}{B\arabic{table}}
\setcounter{table}{0}

\begin{table}[t]

\caption{\textbf{Ablation Results}. Performance comparison of the Qwen2.5-1.5B-instruct model trained on Text-to-SQL collection, Neo4j Text-to-Cypher dataset, or both datasets jointly. The three digits in model names indicate the use of different rewards during training$^*$. We report execution accuracy for SQL tasks (BIRD minidev + evidence, Spider), Exact Match for QA tasks (CRT-QA, CR-LT-KGQA, TableBench), BLEU score for Text2Cypher, and $f1_{exec}$ for EHRSQL.}

\label{tab:appendix_ablations}

\centering
\scriptsize
\begin{tabular}{@{}lccccccc@{}}
\toprule
& \multicolumn{3}{c}{\textbf{Text-to-SQL}} & \multicolumn{2}{c}{\textbf{Tabular QA}} & \textbf{Text-to-Cypher} & \textbf{KG QA} \\
\cmidrule(lr){2-4} \cmidrule(lr){5-6} \cmidrule(lr){7-7} \cmidrule(lr){8-8}
\textbf{Model} & BIRD minidev & Spider & EHRSQL & CRT-QA & TableBench & Text2Cypher & CR-LT-KGQA \\
\midrule

SQL-111 & 11.3\scriptsize{±1.4} & 49.5\scriptsize{±3.1} & \textbf{9.5}\scriptsize{±2.5} & 11.2\scriptsize{±2.1} & 24.0\scriptsize{±2.8} & 9.2\scriptsize{±1.5} & 52.3\scriptsize{±4.6} \\
SQL-011 & 9.5\scriptsize{±1.5} & 46.2\scriptsize{±3.3} & 8.0\scriptsize{±2.6} & 9.7\scriptsize{±2.3} & 20.4\scriptsize{±3.0} & 7.9\scriptsize{±1.4} & 50.2\scriptsize{±4.9} \\
SQL-101 & 10.7\scriptsize{±1.4} & 47.5\scriptsize{±3.2} & 8.7\scriptsize{±2.8} & 10.0\scriptsize{±2.2} & 22.2\scriptsize{±2.9} & 8.7\scriptsize{±1.4} & 51.1\scriptsize{±4.8} \\
SQL-110 & 10.5\scriptsize{±1.6} & 46.7\scriptsize{±3.1} & 8.2\scriptsize{±2.4} & 9.0\scriptsize{±2.1} & 19.3\scriptsize{±2.7} & 8.3\scriptsize{±1.5} & 49.5\scriptsize{±4.5} \\
\midrule
Cypher-111 & 8.2\scriptsize{±2.3} & 45.2\scriptsize{±4.2} & 7.8\scriptsize{±2.5} & 10.2\scriptsize{±2.5} & 22.5\scriptsize{±3.1} & {15.4}\scriptsize{±1.2} & 58.3\scriptsize{±3.9} \\
Cypher-011 & 7.5\scriptsize{±2.4} & 45.1\scriptsize{±4.0} & 7.5\scriptsize{±2.6} & 9.8\scriptsize{±2.3} & 20.1\scriptsize{±3.4} & 13.2\scriptsize{±1.4} & 54.7\scriptsize{±3.8} \\
Cypher-101 & 7.7\scriptsize{±2.5} & 45.5\scriptsize{±4.5} & 7.7\scriptsize{±2.7} & 9.6\scriptsize{±2.4} & 20.6\scriptsize{±3.0} & 14.5\scriptsize{±1.9} & 56.3\scriptsize{±4.2} \\
Cypher-110 & 7.6\scriptsize{±2.3} & 45.3\scriptsize{±4.4} & 7.6\scriptsize{±2.8} & 9.3\scriptsize{±2.4} & 19.7\scriptsize{±3.2} & 13.8\scriptsize{±1.2} & 54.2\scriptsize{±3.7} \\
\midrule
Both-111 & \textbf{12.0}\scriptsize{±1.2} & \textbf{51.2}\scriptsize{±3.3} & {9.4}\scriptsize{±2.4} & \textbf{14.2}\scriptsize{±2.8} & \textbf{30.0}\scriptsize{±3.1} & \textbf{15.8} \scriptsize{±2.2} & \textbf{60.8}\scriptsize{±4.0} \\
Both-011 & 9.8\scriptsize{±1.7} & 48.1\scriptsize{±3.4} & 9.0\scriptsize{±2.6} & 12.0\scriptsize{±2.3} & 26.5\scriptsize{±3.2} & 12.5\scriptsize{±1.3} & 57.0\scriptsize{±4.3} \\
Both-101 & 9.5\scriptsize{±2.3} & 49.3\scriptsize{±3.5} & 9.4\scriptsize{±2.3} & 13.0\scriptsize{±2.3} & 27.3\scriptsize{±3.0} & 13.8\scriptsize{±1.4} & 58.5\scriptsize{±4.1} \\
Both-110 & 9.3\scriptsize{±2.2} & 48.9\scriptsize{±3.6} & 9.1\scriptsize{±2.5} & 11.5\scriptsize{±2.4} & 25.8\scriptsize{±3.5} & 13.2\scriptsize{±1.4} & 56.4\scriptsize{±4.2} \\
\midrule
Qwen2.5-1.5B-instruct & 9.0\scriptsize{±1.6} & 44.7\scriptsize{±3.6} & 7.4\scriptsize{±3.2} & 8.3\scriptsize{±2.3} & 18.0\scriptsize{±2.4} & 7.2\scriptsize{±1.3} & 48.7\scriptsize{±5.5} \\
\bottomrule
\multicolumn{8}{@{}l@{}}{\scriptsize $^*$Model names (e.g., SQL-111) use a three-digit format where each digit (1/0) indicates whether the corresponding }\\
\multicolumn{8}{@{}l@{}}{\scriptsize reward was used during training: string matching (1st digit), LLM reward (2nd digit), and structural reward (3rd digit).}\\
\end{tabular}
\end{table}

\renewcommand{\thetable}{C\arabic{table}}
\setcounter{table}{0}

\section{SM3: Direct Comparison of SQL, Cypher, and MQL} \label{app:sm3}
We evaluate models trained on our Text-to-SQL and Text-to-Cypher datasets using the SM3-Text-to-Query Dataset \cite{sivasubramaniam2024smtexttoquery}, which provides semantically equivalent queries across SQL, Cypher, and MongoDB Query Language (MQL). This allows us to assess how well our training approach generalizes to unseen query languages.  We use the schema information and the few-shot examples from the paper, and use o4 mini as an LLM judge.

Base models demonstrate better performance on SQL compared to Cypher and significantly weaker performance on MQL, suggesting that their behavior reflects the data availability during pretraining (SQL being the most abundant and MQL the scarcest). However, our models outperform in Cypher over SQL, likely due to Cypher's relative simplicity or the greater effectiveness of our structural reward mechanism for graph-based tasks.

Our results in Table~\ref{tab:model_performance} demonstrate strong transfer learning capabilities. Models trained jointly on SQL and Cypher (e.g., Qwen2.5-14B-trained-Both achieving 42.5\% on Cypher and 40.0\% on SQL) outperform their single-task counterparts (Qwen2.5-14B-trained-Cypher achieving 41.8\% on Cypher, Qwen2.5-14B-trained-SQL achieving 39.5\% on SQL). More importantly, this joint training enables better generalization to the completely unseen MQL, with QwQ-32B-trained-Both reaching 20.8\% accuracy compared to base model's 19.4\%. Adding few-shot examples further boosts performance across all settings, with QwQ-32B-trained-Both-fewshot achieving our best results (57.7\% Cypher, 48.8\% SQL, 24.0\% MQL), suggesting effective combination of learned structural understanding and exemplar-based reasoning.

\begin{table}[h]
\caption{\textbf{SM3 - Direct Comparison of SQL, Cypher, and MQL.} Performance on semantically equivalent queries across three query languages using the SM3 benchmark. Each query expresses the same intent in SQL, Cypher, and MongoDB Query Language (MQL), enabling direct comparison of language complexity. LLM execution accuracy is reported. Models fine-tuned on Text-to-SQL are labeled as `trained-SQL', those fine-tuned on Text-to-Cypher as `trained-Cypher', and those trained on both tasks as `trained-Both'. Few-shot results indicate performance when correct Text-to-Query examples are provided in the prompt.}
\label{tab:model_performance}
\centering
\begin{tabular}{@{}lcccccc@{}}
\toprule
& \multicolumn{2}{c}{\textbf{Text-to-Cypher}} & \multicolumn{2}{c}{\textbf{Text-to-SQL}} & \multicolumn{2}{c}{\textbf{Text-to-MQL}} \\
\cmidrule(lr){2-3} \cmidrule(lr){4-5} \cmidrule(lr){6-7}
\textbf{Model}  & Zero-shot & Few-shot & Zero-shot & Few-shot & Zero-shot & Few-shot \\
\midrule
o3 & 40.5±3.9 & \textbf{59.3±3.8} & \textbf{46.3}±3.2 & \textbf{57.7}±3.6 & \textbf{25.7}±4.1 & \textbf{30.3}±3.5 \\
o3-mini & 40.0±5.6 & 53.7±3.3 & 41.3±4.2 & 50.8±3.7 & 15.5±2.5 & 20.0±2.1 \\
\midrule
Qwen2.5-14B-instruct & 32.5±3.2 & 36.7±4.2 & 35.5±3.7 & 41.7±3.8 & 9.0±3.9 & 15.2±2.2 \\
Qwen2.5-14B-trained-SQL & 35.5±2.8 & 38.5±3.8 & 39.5±4.0 & 43.7±3.3 & 9.6±2.2 & 18.2±1.8 \\
Qwen2.5-14B-trained-Cypher & 41.8±3.3 & 49.8±3.9 & 36.7±5.3 & 40.5±4.0 & 10.3±1.2 & 18.5±1.6 \\
Qwen2.5-14B-trained-Both & 42.5±3.2 & 49.9±4.3 & 40.0±4.2 & 44.3±4.3 & 13.8±1.1 & 19.7±2.4 \\
\midrule
Qwen3-14B & 40.5±5.6 & 53.0±4.0 & 38.3±4.8 & 50.7±4.1 & 19.7±2.0 & 19.0±3.5 \\
Qwen3-14B-trained-SQL & 40.7±4.2 & 53.5±3.9 & 41.8±4.0 & \textbf{51.8}±3.3 & 20.0±2.1 & 20.7±1.7 \\
Qwen3-14B-trained-Cypher & 45.2±3.8 & 54.0±4.1 & 40.5±4.7 & 47.5±3.5 & 20.5±2.0 & 20.5±1.9 \\
Qwen3-14B-trained-Both & \textbf{45.7±5.6} & 54.0±4.1 & 42.0±2.2 & 48.0±3.8 & 20.5±2.8 & 20.6±0.4 \\
\midrule
QwQ-32B & 35.8±5.2 & 55.3±4.3 & 41.0±4.2 & 42.4±4.0 & 19.4±2.4 & 20.8±2.7 \\
QwQ-32B-trained-Both & 43.7±5.5 & \textbf{57.7}±4.3 & \textbf{42.5}±4.7 & {48.8}±3.8 & \textbf{20.8}±1.2 & \textbf{24.0}±2.3 \\
\bottomrule
\end{tabular}
\end{table}


\newpage
\section{Prompts}
\label{app:prompts}

This section details the prompts used in our pipeline, closely aligned with \citet{stoisser2025sparks}. For Cypher queries, we simply replace the word "SQL" with "Cypher" in all prompts.

\subsection{Instruction Summarization} \label{app:instruction_summarization}
As part of the synthetic data generation pipeline, we employ a specialized prompt to distill complex natural language SQL task descriptions into concise, structured summaries that capture essential query requirements.

\begin{center}
    \fbox{\parbox{\columnwidth}{\small
    You are a SQL assistant tasked with summarizing instructions for SQL query generation. Below are original natural language task descriptions that outline a specific request. Create a concise summary that highlights the core requirements for the SQL query to be constructed.
    
    \textbf{ORIGINAL INSTRUCTIONS: {original natural language task description}}
    
    When creating the summary, focus on clarity and precision, ensuring that the essential elements necessary for generating the SQL query are retained. You may follow this template:
    
    \textbf{SUMMARY:} Write the summarized instructions here, clearly stating the goals and key aspects required for the SQL query.
    }}
\end{center}

\subsection{Creating Synthetic CoT} \label{app:create_Sythetic_COT}
Our synthetic Chain-of-Thought generation prompt combines schema information, original instructions, and summarized requirements to guide the creation of well-reasoned SQL queries.

\begin{center}
    \fbox{\parbox{\columnwidth}{\small
    You are a SQL expert. Below are SQL table schemas paired with both original and summarized instructions that describe a specific task. Using valid SQLite syntax, write a response that appropriately completes the request for the provided tables.
    
    \textbf{SCHEMA: {schema}}

    \textbf{ORIGINAL INSTRUCTIONS: {original natural language task description}}

    \textbf{SUMMARIZED INSTRUCTIONS: {summarized task instructions}}

    When answering, provide reasoning for the SQL query you create using the following template:
    
    <sql> Write the SQL query here, ensuring it adheres to SQLite syntax and effectively accomplishes the task described in the instructions. </sql>
    }}
\end{center}

\subsection{Evaluation of Synthetic CoT} \label{app:eval_Sythetic_COT}

The evaluation prompt for the synthetic CoT generation pipeline provides a binary assessment framework for comparing generated SQL queries against reference solutions while considering schema constraints.

\begin{center}
    \fbox{\parbox{\columnwidth}{\small
        You are an SQL expert, and your task is to evaluate whether the SQL query below is correct based on the provided schema and the correct SQL reference.
        
        \textbf{SQL Query:} {ans.sql}
        
        \textbf{Schema:} {schema}
        
        \textbf{Correct SQL:} {correct\_sql}
        
        Return ONLY "Correct" or "Wrong".
    }}
\end{center}

\subsection{Training Prompts}
Following \citet{guo2025deepseek}, we use this system prompt that encourages explicit reasoning through Chain-of-Thought.
\begin{center}
    \fbox{\parbox{\columnwidth}{\small
A conversation between User and Assistant. The user asks a question, and the Assistant solves it,
The assistant first thinks about the reasoning process in the mind and then provides the user
with the answer. The reasoning process is enclosed within <think> </think> tags, respectively, i.e., <think> reasoning process here </think> answer here. User: 
    }}
\end{center}

The user prompt provides specific instructions for query generation:

\begin{center}
    \fbox{\parbox{\columnwidth}{\small
        Below is an instruction that describes a task, paired with an input that provides further context. 
        Write a response that appropriately completes the request.
        
        \textbf{Instruction:}
        You are a SQLite expert. Given an input question, create a syntactically correct SQLite query to run. 
        Enclose the final sql query within "```sql" and "```" tags.
        
        \textbf{Input:}
        Here is the relevant table info: \{table\_info\}.

        Write a SQLite query for the following task: \{task\}.
        
        \textbf{Response:}
    }}
\end{center}

\subsection{LLM Evaluation Metric Prompt}

For binary evaluation of query correctness, we employ a prompt that focuses on semantic equivalence while allowing for minor syntactic variations:

\begin{center}
    \fbox{\parbox{\columnwidth}{\small
        You are SQL expert and your task is to evaluate if the predicted SQL query is correct
        based on the Schema and the correct SQL query. If no SQL query was found then the answer is Wrong. 
        The query is considered correct even if the only mistakes are in letter casing (uppercase vs lowercase).
        
        \textbf{Schema:} \{example['context']\}
        
        \textbf{Predicted query:} \{pred\_query\}
        
        \textbf{Correct SQL query:} \{correct\_query\}
        
        Return ONLY "Correct" or "Wrong"
    }}
\end{center}

\subsection{LLM Judge Reward Classification Prompt}
To provide fine-grained feedback during training, we use a five-class classification prompt that assesses query quality across multiple dimensions:

\begin{center}
    \fbox{\parbox{\columnwidth}{\small
        Compare these SQL queries to the correct query and grade each one as: 
        'Very bad', 'Bad', 'Above average', 'Good', or 'Excellent'.
        
        Use the following grading system, with the correct query as reference:
        
        \textbf{Correct Query:} \{true\_query\}
        
        \textbf{1. Excellent:} Perfect match with \{true\_query\}
        
        \textbf{2. Good:} Contains only grammar mistakes
        
        \textbf{3. Above average:} Mostly correct but contains one logical error
        
        \textbf{4. Bad:} Contains multiple mistakes
        
        \textbf{5. Very bad:} No query produced or significantly different from correct query
        
        \textbf{Queries to grade:}
        \{queries\_to\_rank\}
        
        \{format\_instructions\}
    }}
\end{center}

\end{document}